\documentclass[a4paper,12pt]{article}
\usepackage{graphicx,latexsym} 
\usepackage{amsfonts}
\usepackage{latexsym}
\usepackage{amsfonts,amsmath,amsthm,amssymb}
\usepackage{natbib}
\usepackage{url}

\newcommand{\vectornorm}[1]{\left|\left|#1\right|\right|}

\theoremstyle{plain}
\newtheorem{thm}{Theorem}[section]
\theoremstyle{plain}
\newtheorem{lemma}{Lemma}[section]
\theoremstyle{plain}
\newtheorem{prop}{Proposition}[section]
\theoremstyle{plain}

\theoremstyle{plain}
\newtheorem{cor}{Corollary}[section]
\theoremstyle{plain}

\theoremstyle{plain}

\theoremstyle{definition}

\theoremstyle{remark}

\begin{document}
%\begin{frontmatter}
\title{Improving the Johnson-Lindenstrauss Lemma}
\author{\small BY JAVIER ROJO AND TUAN S. NGUYEN}%$^1$  
\date{}
\maketitle{}
%\vspace{-5cm}
\begin{quote}
%\hspace{1cm}
\centerline{\it Rice University \\
$jrojo@rice.edu$\\
Phone: 713-348-2797
}

\end{quote}
%\author{Javier Rojo\corref{cor1}\fnref{label}}
%\footnote{Typeset names in 8~pt roman,  
%uppercase. Use the footnote to indicate the
%present or permanent address of the author.}}
%\ead{jrojo@rice.edu}
%\author {Tuan S. Nguyen\fnref{label}}
%\ead{tsn4867@rice.edu}
%\address{${}^1$Statistics Department, MS 138, Rice University, 6100 Main Street,
%Houston, Texas 77005\\
%Phone: 713-348-2797 and 713-348-3658}

%\numberwithin{equation}{section}

%\begin{abstract}
\begin{quote}
The Johnson-Lindenstrauss Lemma allows for the projection of $n$ points in  $p-$dimensional Euclidean space onto a $k-$dimensional Euclidean space, with $k \ge \frac{24\ln \emph{n}}{3\epsilon^2-2\epsilon^3}$, so that the pairwise distances are preserved within a factor of $1\pm\epsilon$. Here, working directly with the distributions of the random distances rather than resorting to the moment generating function technique, an improvement on the lower bound for $k$ is obtained. The additional reduction in dimension when compared to bounds found in the literature, is at least $13\%$, and, in some cases, up to $30\%$ additional reduction is achieved. Using the moment generating function technique, we further provide a lower bound for $k$ using pairwise $L_2$ distances in the space of points to be projected and pairwise $L_1$ distances in the space of the projected points. Comparison with the results obtained in the literature  shows that the bound presented here provides an additional $36-40\%$  reduction. \\
%\end{abstract}
\end{quote}

%\begin{keyword}
%Embedding  \sep Gaussian random matrix \sep dimension reduction.
%\end{keyword}

%\end{frontmatter}

\section{Introduction}
With the arrival of the ``small $n$, large $p$'' paradigm, dimension reduction methods have come to the forefront of many applications. For example, in survival analysis studies using microarray data, in the order of 10-20K expressions per patient are collected. On the other hand, usually only  a few hundred patients are available for the study. For this reason, one must reduce the dimension of the gene expression data matrix, before embarking on any type of analysis. The challenge of ``small $n$, large $p$'' arises also in high throughput molecular screening, astronomy, and image analysis. Among the various dimension reduction techniques that are used -- some new, some old -- the Random Projection method has attracted a lot of attention lately. Random Projection (RP) provides a computational method for dimension reduction whereby the original $p-$dimensional data points are projected onto a \linebreak $k-$dimensional subspace by multiplying the $n$ x $p$ data matrix $X$ by a $p$ x $k$ random matrix $\Gamma$. 
In matrix notation, 
\begin{equation}
\label{RP1a}
T=X\Gamma
\end{equation}
where $X$ is the $n $ x $p$ data matrix, $\Gamma$ is a $p $ x $ k$ random matrix, and $T$ is the resulting $n $ x $k$ matrix consisting of the projected points onto a lower $k-$dimensional subspace.  

Orthogonality of the projection matrix preserves similarities, e.g. the inner product or Euclidean distance, of the original vectors when projected to the low-dimensional space. Although the random matrix $\Gamma$ is not orthogonal,
Achlioptas \cite{Achlioptas} pointed out that  the loss of information is minimal because the orthogonal property is achieved with high probability in high-dimensional space. %The unitary property can be avoided since in high-dimensional space, the length of each of the row vectors of $\Gamma$ is concentrated around $k$ \cite{Matousek2}. That is, $E(r_{i1}^2 + \dots + r_{ik}^2) =k$, where $i$ denotes the $i^{th}$ row of $\Gamma$.  
Moreover, the random projection matrix $\Gamma$ is  close to orthogonal in high-dimensional space, i.e. $\frac{1}{k}\Gamma \Gamma^T \approx I_p$ \cite{Hecht_Nielsen}.
%, because the number of almost orthogonal directions is much larger than the number of orthogonal directions in high-dimensional space.%
Using Random Projections, orthogonalization of the projection matrix in high-dimensional space can be avoided without losing much information in the original data \cite{Achlioptas,Goel}. 

Random Projection (RP) has been used in numerous areas such as in machine learning \cite{Arriaga,Bingham,Candes,Dasgupta2,FB,Fradkin,Kaski}, latent semantic indexing \cite{Papadimitriou,Kurimo,Vempala}, face recognition \cite{Goel}, kernel computations \cite{Achlioptas2,Blum}, nearest neighbor queries \cite{Deegalla,Kleinberg,Indyk2}, privacy preserving distributed data mining \cite{Liu}, gene expression clustering \cite{Bertoni1,Bertoni2,Bertoni3}, and finding DNA motifs \cite{Buhler}. A good overview on the use of RP is given in \cite{Bingham,Goel}. While several dimension reduction methods obtain the low-dimensional subspace by optimizing a certain criteria, RP does not. For example, Principal Component Analysis (PCA) finds the set of directions that maximize the variance in the data. It turns out that the performance of RP is comparable to PCA in face recognition experiments \cite{Goel}, text and image data \cite{Bingham}, and machine learning \cite{Fradkin}. In addition, although PCA is a popular dimension reduction method, it is quite expensive computationally since it involves computing the eigenvalue decomposition of the data covariance matrix \cite{Mannila}. RP, on the other hand, is simple and computationally efficient. The computing cost for PCA is $O(np^2)+O(p^3)$, while that of
RP is $O(k^2p)$ when the entries to the random matrix are independent and identically distributed (i.i.d.) standard Gaussians, and $O(kp)$ when the entries are of Achlioptas type \eqref{ach1} \cite{Goel,Li4,Bingham}. 
%The algorithmic applications of RP is discussed in \cite{Indyk}.  

The main motivation for Random Projection (RP) is the Johnson-Lindenstrauss Lemma (1984), which states that a set of \emph{n} points in $p-$dimensional 
Euclidean space can be mapped down onto a $k=O(\ln n/{\epsilon^2})$ 
dimensional Euclidean space such that the pairwise distance between any two points 
is preserved within a factor of $(1 \pm \epsilon)$ for any $0 < \epsilon < 1$. We should note that the similarity measure used in the JL Lemma is the Euclidean distance. In the original proof of the JL Lemma, Johnson and Lindenstrauss \citep{JL} show that such a mapping is provided by a random orthogonal projection. Frankl and Maehara \citep{Frankl1}  simplified the original proof of Johnson and Lindenstrauss using geometric techniques, and provided an improvement on the lower bound for $k$, i.e. $k\ge \left \lceil \frac{27\ln \emph{n}}{3\epsilon^2-2\epsilon^3} \right \rceil + 1$. Indyk and Motwani \citep{Indyk2} simplified the proof of the JL Lemma using i.i.d. standard Gaussian entries for the random matrix $\Gamma$. Also, using a Gaussian random matrix, Dasgupta and Gupta \citep{Dasgupta} further simplified the proof with elementary probabilistic techniques based on moment generating functions, and improved on the lower bound for $k$ to be $k\ge \frac{24\ln  n}{3\epsilon^2-2\epsilon^3}$.  
%Several %probabilistic% 
%versions of the JL Lemma were proposed in the literature (for example, see Frankl and Maehara \citep{Frankl1}, Indyk and Motwani \citep{Indyk2}, Dasgupta and Gupta \citep{Dasgupta}, and Matousek \citep{Matousek}). %For the probabilistic versions of the JL Lemma, the entries of $\Gamma$, denoted by $r_{ij}$ where $i=1,...,p$ and $j=1,...,k$ with $k \ll min(n,p)$, are independent and identically distributed (i.i.d.) with zero mean. 
%This is the only necessary condition for preserving pairwise distances \citep{Arriaga}. 
%It is typical to let $r_{ij}$ to be i.i.d. with zero mean and unit variance. 
%A convenient choice for the random matrix $\Gamma$ is the random matrix with i.i.d. $N(0,1)$ entries, as used in the proof of the probabilistic version of the JL Lemma given by \citep{Dasgupta}. However,
 
Instead of improving the lower bound for $k$, several papers in the literature focus on improving the computational time of the Random Projections. %Using the moment generating function approach, they obtained the same lower bound for $k$ as in Dasgupta and Gupta \citep{Dasgupta}.
Achlioptas \citep{Achlioptas} proposed two simpler distributions as alternatives to using a Gaussian random matrix:\\
\begin{equation}
\label{ach1}
	r_{ij}=\begin{cases} 
				+1& \text{with prob. 1/2} \\ 
				-1&  \text{with prob. 1/2}
		\end{cases}  
\end{equation}
and
\begin{equation}
\label{ach2}
	r_{ij}= \sqrt{3}\begin{cases} 
				+1& \text{with prob. 1/6} \\
				0& \text{with prob. 2/3}\\
				-1&  \text{with prob. 1/6} \text{ .}
		\end{cases}  
\end{equation}
These distributions are easy to implement, and the computational time is greatly
reduced. With the distribution in \eqref{ach1}, only 1/2 of the operations are needed which
implies a 2-fold speedup. Similarly, a 3-fold speedup is obtained for the distribution in \eqref{ach2}. The random matrix defined through  \eqref{ach1} and \eqref{ach2} can be generalized as follows:
\begin{equation}
\label{ach3}
	r_{ij}= \sqrt{q}\begin{cases} 
				+1& \text{with prob.} \quad \frac{1}{2q} \\
				0& \text{with prob.} \quad 1-\frac{1}{q}\\
				-1&  \text{with prob.} \quad \frac{1}{2q} \text{ .}
		\end{cases}  
\end{equation}
Thus, $q = 1$ yields \eqref{ach1}, and $q = 3$ yields \eqref{ach2}. Furthermore,
using $q \gg 3$ (e.g. $q = \sqrt{p}$ or $q = \frac{p}{\ln p}$) can 
significantly speed up the computation \citep{Li4} since the random matrix $\Gamma$ is \emph{very sparse}. Ailon and Chazelle \citep{Ailon} extended the idea of using sparse random matrices with a randomized Fourier transform to speed up the RP. The running time of the Ailon and Chazelle algorithm is improved by using a 4-wise independent deterministic code matrix with randomized block diagonal matrix \citep{Ailon2}, and by using any deterministic matrix with tensor products and Lean Walsh Transform \citep{Ailon3}. Matousek \citep{Matousek} provided a version of the JL Lemma that allows the entries of the random matrix $\Gamma$ to be arbitrary independent random variables with zero mean, unit variance and subgaussian tail  (see \citep{Matousek} for a discussion on the variants of the JL Lemma). All these improvements on the time needed to obtain the random projection, however, do not improve on the lower bound for $k$. 

We adopt the following notation to use throughout the paper. Denote by $\phi(.)$ and $\Phi(.)$ the standard Gaussian density and cumulative distribution functions, respectively. Denote by $L_2$-$L_2$ RP the random projection that uses $L_2$ distances in the space of points to be projected and $L_2$ distances in the space of the projected points, and $L_2$-$L_1$ RP the random projection that uses $L_2$ distances in the space of points to be projected and $L_1$ distances in the space of the projected points.  For $\mathbf{x}\in \mathbf{R}^p$, let $\vectornorm{\mathbf{x}}_1 = \sum_{i=1}^p |x_i|$,  and $\vectornorm{\mathbf{x}}^2 = \sum_{i=1}^p x_i^2$. 

In this paper, the JL Lemma  is revisited. In particular, an improvement on the lower bound for $k$ from Dasgupta and Gupta \citep{Dasgupta} is provided by working directly with the exact distribution of the random Euclidean distances rather than using the moment generating function approach. The additional reduction provided by our results is at least $13\%$ in all cases, and in other cases an additional reduction of $30\%$ is possible when compared to the bound obtained in Dasgupta and Gupta \citep{Dasgupta}. The JL Lemma uses Euclidean distance to  measure  the distortion in the distances of the projected points when projecting from
$p-$dimensional Euclidean space onto $k-$dimensional Euclidean space. This paper also obtains a lower bound for $k$ for the $L_2$-$L_1$ RP.  
A lower bound for $k$ using random matrices as in \eqref{ach3} with $q=1,2,3$ is also provided for the $L_2$-$L_1$ RP. This improved lower bound provides an additional $36-40\%$ reduction in dimension when compared to the results obtained in Matousek \citep{Matousek}.

The paper is organized as follows: section 2 discusses the JL Lemma in detail. The improvement to the JL bound for the $L_2$-$L_2$ RP is discussed in section 3. A lower bound for $k$ using $L_2$-$L_1$ RP is provided in section 4, and section 5 provides some concluding remarks. Most of the technical proofs are relegated to an appendix.

\section{Johnson-Lindenstrauss  Lemma}
%\subsection{Applications of RP}

%In extending the Lipschitz maps into a Hilbert space, Johnson and Lindenstrauss \citep{JL} discovered a mathematical fact, from which the JL Lemma took its name.  
In their pioneering work, Johnson and Lindenstrauss \citep{JL} provided the following result:

\noindent\textbf{Johnson-Lindenstrauss (JL) Lemma} \textit{For any} $0 < \epsilon < 1$ \textit{and integer} $n$\textit{, let} $k$
	\textit{be such that} 
%\begin{equation}
%\label{eq:JL1}
	 $k = O(\ln n/{\epsilon^2})$. 
%\end{equation}
\textit{For any set V of $n$ points in} ${\mathbf R}^p$, \textit{there is a linear map} $f\colon{\mathbf R}^p \rightarrow {\mathbf R}^k$
	\textit{such that for any} $\mathbf{u},\mathbf{v} \in V,$ 
\begin{equation}
\label{eq:JL2}
	(1-\epsilon) \vectornorm{\mathbf{u}-\mathbf{v}}^2 \le 
	\vectornorm{\mathbf{f(u)}-\mathbf{f(v)}}^2 \le
	(1+\epsilon) \vectornorm{\mathbf{u}-\mathbf{v}}^2 \text{ .}
\end{equation}

%\vspace{1.5pc}

Johnson and Lindenstrauss \citep{JL} showed that the linear map $f$ can be taken to be a random orthogonal projection, but the
explicit construction of $f$ is not provided. Indyk and Motwani \citep{Indyk2} and Dasgupta and Gupta \citep{Dasgupta} gave an explicit form of the mapping $f$ in their versions of the JL Lemma. The mapping is provided by $f(\mathbf{x})=\mathbf{x} \Gamma$, where $\mathbf{x}\in V$ and where entries of the random matrix $\Gamma$ are i.i.d. standard Gaussians. In a remarkable paper using only elementary probabilistic techniques, Dasgupta and Gupta \citep{Dasgupta} improved on the lower bound for $k$ from the original JL Lemma as follows. 

%The lower bound for $k$ is improved by Dasgupta and Gupta from the original JL Lemma as follows.

\noindent\textbf{ Dasgupta and Gupta version of the JL Lemma: } 
\textit{For any} $0 < \epsilon < 1$ \textit{and integer} $n$\textit{, let} $k$
	\textit{be such that} 
\begin{equation*}
\label{eq:JL1}
	 k \ge \frac{24 \ln \emph{n}}{3\epsilon^2-2\epsilon^3}\text{ .}
\end{equation*}
\textit{For any set V of n points in} ${\mathbf R}^p$, \textit{there is a linear map} $f\colon{\mathbf R}^p \rightarrow {\mathbf R}^k$
	\textit{such that for any} $\mathbf{u},\mathbf{v} \in V,$ 
\begin{equation}
\label{eq:JL3b2}
P\left[(1-\epsilon) \vectornorm{\mathbf{u}-\mathbf{v}}^2 \le 
	\vectornorm{\mathbf{f(u)}-\mathbf{f(v)}}^2 \le
	(1+\epsilon) \vectornorm{\mathbf{u}-\mathbf{v}}^2\right] \ge 1-\frac{2}{n^{2}} \text{ .}
\end{equation}
%For each of the ${n}\choose{2}$ pairs $\mathbf{u},\mathbf{v} \in V$, the Euclidean  
%distance is preserved within a factor of $1\pm \epsilon$ when projecting from high to low-dimensional Euclidean space with probability at least $1-2/{n^{2+\beta}}$. Hence, the probability of not obtaining such a projection is bounded by ${{n}\choose{2}} \left(2/{n^{2+\beta}}\right) <1/n^{\beta}$.  

Let $\mathbf{x}=\mathbf{u}-\mathbf{v}$. Since $f$ is linear, the inequality in \eqref{eq:JL3b2} is equivalent to 
\begin{equation}
\label{eq:JL3b3}
P\left[\vectornorm{\mathbf{f(x)}}^2 \ge (1+\epsilon) \vectornorm{\mathbf{x}}^2\right] + P\left[\vectornorm{\mathbf{f(x)}}^2 \le (1-\epsilon) \vectornorm{\mathbf{x}}^2\right]\le \frac{2}{n^{2}} \text{ .}
\end{equation}

The bound in \eqref{eq:JL3b3} can be obtained by separately bounding the left- and right-tail probabilities. That is, by finding $f$ so that simultaneously, 
\begin{equation}
\label{eq:JL3b4}
P\left[\vectornorm{\mathbf{f(x)}}^2 \ge (1+\epsilon) \vectornorm{\mathbf{x}}^2\right] \le \frac{1}{n^{2}} \text{ ,}
\end{equation}
and
\begin{equation}
\label{eq:JL3b42}
P\left[\vectornorm{\mathbf{f(x)}}^2 \le (1-\epsilon) \vectornorm{\mathbf{x}}^2\right] \le \frac{1}{n^{2}} \text{ .}
\end{equation}

The proof of Dasgupta and Gupta's version of the JL Lemma hinges on the use of standard Gaussians as entries to the random matrix $\Gamma$, and the moment generating function technique. The proof is sketched next, as this will set down the notation and facilitate the reading of section 3. 

\noindent\textbf{Sketch of the proof of Dasgupta and Gupta's version of the JL Lemma} \\
\noindent     
Let $\Gamma$ to be a random matrix of 
	dimension \textit{p} x \textit{k} with entries $r_{ij} \sim N(0,1)$ independent. For $\mathbf{x}\in V$, define $f(\mathbf{x})= \frac{1}{\sqrt{k}}\mathbf{x}\Gamma$, and  $\mathbf{y} =\sqrt{k} \frac{f(\mathbf{x})}
{\vectornorm{\mathbf{x}}}$. Then $y_j = \frac{\mathbf{x}r_j}{\vectornorm{\mathbf{x}}}\sim N(0,1)$ and $y_j^2 \sim \chi_1^2$ with $E(\vectornorm{\mathbf{y}}^2) = k$, where $r_j$ is the $j^{th}$ column of $\Gamma$. 	
%Now we need to show:
%\begin{eqnarray}
%\label{JL4b}
% \text{If \emph {k =  {O(log n/$\epsilon^2$)}}, then} 
% \vectornorm{\mathbf{f(x)}}^2 &\approx(1\pm\epsilon)\vectornorm{\mathbf{x}}^2k,\nonumber\\
%&\text{ with probability} \ge 1-\frac{2}{n^2}. 
%\end{eqnarray}
%From this, the result of the Johnson-Lindenstrauss lemma follows because of 
%the linearity of $f$.

Let $\alpha_1 = k(1+\epsilon)$, $\alpha_2 = k(1-\epsilon)$. 
Then the right-tail probability is bounded by
\begin{eqnarray}
\label{JL4c}
 P\left[\vectornorm{\mathbf{f(x)}}^2 \ge (1+\epsilon) \vectornorm{\mathbf{x}}^2\right] 
 &=& P\left[ \vectornorm{\mathbf{y}}^2 \ge k(1+\epsilon) \right] \\
 %&= & P[\underbrace{e^{s\vectornorm{\mathbf{y}}^2}e^{-s\alpha_1}}_{> 0}\ge 1] \nonumber\qquad \forall s\ge 0\\
% &\le & E\left[e^{s\vectornorm{\mathbf{y}}^2}e^{-s\alpha_1} \right] \nonumber
% \qquad \textrm{(Markov's Inequality, } s>0)\\
% &=& e^{-s\alpha_1}E\left[e^{s \sum_{j=1}^k y_j^2} \right] \nonumber\\
 %&=& e^{-s\alpha_1}E\left(\prod_{j=1}^k e^{s y_j^2} \right)
  \label{JL4cb11}
&\le& \left( e^{-s(1+\epsilon )}E(e^{s y_j^2}) \right)^k \text{ ,}
  \qquad s>0  \\
  \label{JL4cb2}
  &=& e^{-s\alpha_1}(1-2s)^{-k/2} \text{ ,}\qquad s\in(0,1/2) 
\end{eqnarray}
where the inequality in \eqref{JL4cb11} follows from Markov's inequality and the fact that the $y_j$'s are i.i.d.
Similarly, the left-tail probability is bounded by
\begin{eqnarray}
\label{JL4c2}
P\left[ \vectornorm{\mathbf{f(x)}}^2 \le (1-\epsilon )\vectornorm{\mathbf{x}}^2 \right] &=&   P[\vectornorm{\mathbf{y}}^2 \le k(1-\epsilon ) ]\\
&\le & \left( e^{s(1-\epsilon )}E(e^{-s y_j^2}) \right)^k \text{ ,} \qquad s>0\\
%&\le & \left( e^{s(1-\epsilon )}(1-sE(y_j^2) + \frac{s^2}{2}E(y_j^4)) \right)^k\nonumber\\
%\label{JL4c3}
%&=& e^{s\alpha_2} \left( 1-s+s^2\right)^k\\
\label{JL4c4}
&\le & e^{-s\alpha_1}(1-2s)^{-k/2} \text{ ,}\qquad s\in(0,1/2) %\text{ .}
\end{eqnarray}
where the inequality in \eqref{JL4c4} follows from the fact that $e^s/{(1+2s)}$ is decreasing in $s\in \left(-\frac{1}{2},\frac{1}{2}\right)$, and hence $\frac{e^{s(1-\epsilon)}}{1+2s}\le \frac{e^{-s(1+\epsilon)}}{1-2s}$ for $s\in(0,1/2)$. %Essentially, the bound for the left-tail probability is the same as that for the right-tail probability. 
The tightest bound in \eqref{JL4cb2}, and hence in \eqref{JL4c4} also, is obtained by minimizing with respect to $s$. The minimizing $s^*=\frac{1}{2}\left(\frac{\epsilon}{1+\epsilon}\right)\in(0,1/2)$. Since 
%\begin{equation}
$g(s) = e^{-s(1+\epsilon )}(1-2s)^{-1/2} \text{ ,  }s\in (0,1/2)$, is strictly convex,
%\end{equation}
%then $g''(s) = \frac{e^{-s(1+\epsilon )}}{(1-2s)^{5/2}} \left(
%2s^2(1+\epsilon)^2 - 4s\epsilon(1+\epsilon)+(2+\epsilon^2)\right) > 0$. This 
$s^*$ is the unique minimizer of \eqref{JL4cb2}. 
Plugging $s^*$ back into \eqref{JL4cb2} yields
\begin{eqnarray}
\label{key2b}
P[\vectornorm{\mathbf{y}}^2 \ge \alpha_1 ] 
&\le & exp\left( -\frac{k}{2} \left(\epsilon - \ln{(1+\epsilon)} \right) \right)\\
\label{key2b2}
&\le & exp\left(-\frac{k}{12} \left(3\epsilon^2-2\epsilon^3\right)\right) \text{ ,}
\end{eqnarray}
where \eqref{key2b2} is obtained after using the inequality $\ln(1+\epsilon)\le \epsilon - \frac{\epsilon^2}{2} + \frac{\epsilon^3}{3}$. 

The same bound is obtained for the left-tail probability. Thus, when \linebreak 
$k \ge \frac{24\ln \emph{n}}{3\epsilon^2-2\epsilon^3}$, then both 
$P\left[ \vectornorm{\mathbf{f(x)}}^2 \ge (1+\epsilon )\vectornorm{\mathbf{x}}^2 \right]$
and
$P\left[ \vectornorm{\mathbf{f(x)}}^2 \le (1-\epsilon )\vectornorm{\mathbf{x}}^2\right]$ are bounded by $1/n^{2}$.

%---------------------------------------------------------%
\section{Improvement on the bound provided by the JL Lemma.} In the previous proof, the left-
and right-tail
probabilities are bounded by using Markov's inequality. The bound can be improved by working directly with the exact probability distribution of the random Euclidean distances. The following Lemma (proof is in the Appendix) is key to proving the main result of this section.

\begin{lemma} \label{lem_g} Let $k$ be an even integer, and $0<\epsilon<1$. Let $\lambda_1 = k(1+\epsilon)/2$ and $d=k/2$. Then 
\begin{equation}
\label{eq:FS2}
	 g(k,\epsilon ) = 
e^{-\lambda_1} \frac{\lambda_1^{d-1}}{(d-1)!}
\end{equation}
is a decreasing function in $k$.
\end{lemma}

\setcounter{thm}{1}
The lower bound for $k$ can then be obtained from the following Theorem.  
\begin{thm} \label{thm1}
For any $0<\epsilon<1$ and integer $n$, let $k$ be the smallest even integer satisfying $\left(\frac{1+\epsilon}{\epsilon}\right) g(k,\epsilon ) \le \frac{1}{n^{2}}$.
Then, for any set V of $n$ points in ${\mathbf R}^p$, there is a linear map $f\colon{\mathbf R}^p \rightarrow {\mathbf R}^k$ such that for any $\mathbf{u},\mathbf{v} \in V,$ 
\begin{equation}
\label{eq:FS3}
P\left[(1-\epsilon) \vectornorm{\mathbf{u}-\mathbf{v}}^2 \le 
	\vectornorm{\mathbf{f(u)}-\mathbf{f(v)}}^2 \le
	(1+\epsilon) \vectornorm{\mathbf{u}-\mathbf{v}}^2\right] \ge 1-\frac{2}{n^{2}}\text{ .}
\end{equation}
\end{thm}

The lower bound for $k$ can be obtained numerically by finding the smallest even integer $k$ satisfying the inequality $\left(\frac{1+\epsilon}{\epsilon}\right) g(k,\epsilon ) \le \frac{1}{n^{2}}$.

Next, we provide the proof to Theorem \ref{thm1}. 

\noindent\textbf{Proof of Theorem \ref{thm1}:}   
Recall the well-known 
\noindent {Gamma-Poisson Relationship: } 
\textit{Suppose} $X \sim Gamma(d, 1)$\textit{, and }$Y \sim Poisson(x)$\textit{. Then
we have }$P(X \ge x) = P(Y \le d - 1)$\textit{. That is,} 
\begin{equation}
\label{gamma1}
\int_x^{\infty} \frac{1}{\Gamma (d)} z^{d - 1} e^{-z} dz
= \sum_{y=0}^{d - 1} \frac{x^y e^{-x}}{y!} 
\end{equation}  
\textit{for }$d = 1,2,3,\dots$. 

Since $\vectornorm{\mathbf{y}}^2 = \sum_{j=1}^{k} y_j^2 \sim \chi_{k}^2 \overset {D}{=} Gamma(k/2,2)$, using \eqref{gamma1} with $\alpha_1 = k(1+\epsilon)$, and setting $d=k/2$, the right-tail probability can be written as,
%Now, let $W = \sum_{j=1}^k y_j^2$, where $y_j$ is given in \eqref{yj}, then $W \sim \chi_k^2$. The pdf of $W$ is given by
%\begin{equation}
%\label{gamma2}
%f(w) = \frac{(1/2)^{k/2}}{\Gamma(k/2)} w^{k/2 - 1} e^{-w/2}
%\end{equation}
%and the cdf is given by:
%\begin{equation}
%\label{gamma3}
%F(x) = P(W \le x) = \int_0^x \frac{(1/2)^{k/2}}{\Gamma(k/2)} w^{k/2 - 1} e^{-w/2} dw \text{ .}
%\end{equation}
%Instead of using Markov inequality and resorting to the moment-generating function
%for the $\chi_k^2$ distribution to establish the lower bound as in Dasgupta's proof, 
%we can  work directly with the cdf of the $\chi_k^2$ random variable.
 %By letting $z = w/2$, we have $2dz = dw$, and letting $a = k/2$, equation \eqref{gamma3}
%becomes:
%\begin{eqnarray}
%\label{gamma4}
%F(x) &=& \int_0^{x/2} 2\left(\frac{(1/2)^a}{\Gamma(a)}\right) (2z)^{a - 1} e^{-z} dz\nonumber\\
%	&=& \int_0^{x/2} \frac{1}{\Gamma(a)}(1/2)^a(2^a) z^{a-1} e^{-z} dz\nonumber\\
%	&=& \int_0^{x/2} \frac{1}{\Gamma(a)} z^{a-1} e^{-z} dz
%\end{eqnarray}
%By Theorem \ref{thm1}, the right-tail probability can be written as, 
\begin{equation}
\label{gamma5}
P[\vectornorm{\mathbf{y}}^2 \ge \alpha_1 ] 
	= e^{-\alpha_1 / 2} \sum_{y=0}^{d-1} \frac{(\alpha_1 /2)^y }{y!}\nonumber \text{ ,}	
\end{equation}
and with $\alpha_2 = k(1-\epsilon)$,
%with $\lambda_2 = \alpha_2 /2 = k(1-\epsilon)/2$, 
the left-tail probability can be written as,
\begin{equation}
\label{gamma7}
P(\vectornorm{\mathbf{y}}^2  \le \alpha_2) = e^{-\alpha_2 / 2}\sum_{y=d}^{\infty} \frac{(\alpha_2 /2)^y }{y!}\text{  .}
\end{equation}

We introduce the following Theorem (proof is in the Appendix), which is essential in 
establishing the bound for the tail probabilities. 

\begin{thm} \label{thm1_5} Let $d$ be a positive integer.\\
a) Let $1 \le d < \lambda_1$. Then,
\begin{equation}
\label{rm2ob}
\sum_{y=0}^{d-1} \frac{\lambda_1^y}{y!} \le  \left(\frac{\lambda_1}{\lambda_1 -d}\right)
\left(\frac{\lambda_1^{d-1}}{(d-1)!} \right) \text{  .}
\end{equation}

\noindent b) Let $0 < \lambda_2 < d$. Then,
\begin{equation}
\label{rm2od}
\sum_{y=d}^{\infty} \frac{\lambda_2^y}{y!} \le  \left(\frac{\lambda_2}{d -\lambda_2}\right)
\left(\frac{\lambda_2^{d-1}}{(d-1)!} \right) \text{  .}
\end{equation}
\end{thm}

Using Theorem \ref{thm1_5}, with $\lambda_1 = \alpha_1/2 = k(1+\epsilon)/2$ and $d=k/2$, the right-tail probability is bounded as follows
\begin{equation}
\label{rm2l}
P[\vectornorm{\mathbf{y}}^2\ge \alpha_1 ] = e^{-\lambda_1}\sum_{y=0}^{d-1} \frac{\lambda_1^y}{y!} \le 
\left(\frac{1+\epsilon}{\epsilon}\right) \left(\frac{\lambda_1^{d-1}}{(d-1)!}\right)e^{-\lambda_1} \text{ .}
\end{equation} 

For the left-tail probability, setting 
$\lambda_2=\alpha_2/2 = k(1-\epsilon)/2$, it follows from Theorem \ref{thm1_5} that
\begin{eqnarray}
\label{rm2l2}
P[\vectornorm{\mathbf{y}}^2\le \alpha_2 ] &=& e^{-\lambda_2}\sum_{y=d}^{\infty} \frac{\lambda_2^y}{y!} \\
\label{rm2l2b}
&\le & \left(\frac{1-\epsilon}{\epsilon}\right) \left(\frac{\lambda_2^{d-1}}{(d-1)!}\right)e^{-\lambda_2}\\
\label{rm2l2c}
&\le & \left(\frac{1+\epsilon}{\epsilon}\right) \left(\frac{\lambda_1^{d-1}}{(d-1)!}\right)e^{-\lambda_1}
\end{eqnarray} 
where the last inequality follows since $e^{\lambda_1-\lambda_2} \le \left(\frac{\lambda_1}{\lambda_2}\right)^d$. Note that the bound for the left-tail probability is the same as that for the right-tail probability. Thus, 
\begin{equation}
\label{rm2l2cb}
P[\vectornorm{\mathbf{y}}^2\ge \alpha_1 ] + P[\vectornorm{\mathbf{y}}^2\le \alpha_2 ]\le 2 \left(\frac{1+\epsilon}{\epsilon}\right) 
g(k,\epsilon)
\end{equation}

For a given $\epsilon$, we can obtain the lower bound for $k$ by numerically obtaining the smallest even integer $k$ such that $\left(\frac{1+\epsilon}{\epsilon}\right) g(k,\epsilon )$ is less than or equal to$\text{ }1/{n^{2}}$. \hspace{3pc}$\Box$\\

A numerical comparison of the two bounds is presented in Table \ref{table:beta} and will be discussed in detail in section 5.

The Johnson-Lindenstrauss (JL) Lemma states that a set of $n$ points in any Euclidean space can be mapped to a Euclidean space of dimension $k=O(\ln n/{\epsilon^2})$ such that the pairwise distance between the points are preserved within a factor of $1\pm \epsilon$. 
 Since the $L_1$ distance is more robust against outliers than the $L_2$ distance, it is of interest to explore the effect of Random Projection on dimension reduction using the $L_1$ norm. In other words, a linear mapping for a set of $n$ points from $p-$dimensional space to $k=O(\ln n/{\epsilon^2})$ dimensional  space is desirable so that the pairwise $L_1$ distances between the points are preserved within a factor of $1\pm \epsilon$. 
However, due to the results of Brinkman and Charikar \citep{Brinkman}, Charikar and Sahai \citep{Charikar}, Lee and Naor \citep{Lee}, and Indyk \citep{Indyk1}, the JL Lemma cannot be extended to the $L_1$ norm using a linear mapping. %In particular, points in $R^p$ cannot be isometrically flattened to a logarithmic dimension \citep{Matousek} using $L_1$ norm.  
Li \textit{et al.} \citep{Li5} proposed three nonlinear mappings (bias-corrected sample median, bias-corrected geometric mean, and bias-corrected maximum likelihood mappings) using $L_1$ norm with standard Cauchy as entries to the random matrix, and obtained $k=O(\ln n/{\epsilon^2})$.

Although it is not possible in the case of a linear mapping to obtain a totally satisfying result when the $L_1$ norm is used to measure distances in both the space of points to be projected and the space of projected points, it is possible to obtain good results by using the $L_2$ norm in the space of points to be projected and the $L_1$ norm to measure distance between the projected points, as discussed next.

%%%%%%%%%%%%%%%%%%%%%%%%%%%%%%%%
\section{ $L_2$--$L_1$ norm with Gaussian Random Matrix} Here a theorem for the linear projection of $n$ points in $p-$dimensional space onto a $k-$dimensional space using i.i.d. standard Gaussians as entries of the random matrix $\Gamma$ 
is presented where the $L_2$ norm is used as a distance in the original space, and $L_1$ is used as a distance in the $k-$dimensional target space. It turns out that 
the original $L_2$ pairwise distances are within a factor of $(1\pm \epsilon) \sqrt{2/{\pi}}$ of the projected $L_1$ distances. For the same factor of $(1\pm \epsilon) \sqrt{2/{\pi}}$, Ailon and Chazelle \citep{Ailon} (sparse Gaussian random matrix with fast Fourier transform) and Matousek \citep{Matousek} (sparse Achlioptas-typed random matrix) obtain the lower bound for $k$ to be:
\begin{equation}
\label{L2_L1a1}
k\ge C\epsilon^{-2} (2\ln (1/{\delta}))
\end{equation}
where $\delta\in (0,1)$, $\epsilon \in (0,1/2)$, and $C$ is a sufficiently large constant. Here, $\delta$ is a parameter that relates to the probability with which any two projected points remain within $(1\pm \epsilon) \sqrt{2/{\pi}}$ of the $L_2$ distance of the original points. Although the multiplicative constant $C$ is not provided, it was taken to be $1$ in one of the proofs in \citep{Matousek}. When $\delta = 1/n^2$, then $k = O\left(\frac{4\ln n}{\epsilon^2}\right)$. 

The following Theorem gives an improvement on the lower bound for $k$ provided by Ailon and Chazelle \citep{Ailon} and Matousek \citep{Matousek}. 

In what follows, for $s>0$, let $A(s) = 2 e^{-s\sqrt{2/{\pi}}(1+\epsilon ) +s^2/2} \Phi(s)$. For a given $\epsilon \in (0,1)$, let $s^*(\epsilon)$ be the value that minimizes $A(s)$. Equivalently, let $s^*$ be the unique solution to 
$s = \sqrt{2/{\pi}} (1+\epsilon) - \frac{\phi(s)}{\Phi(s)}$. 

\begin{thm} \label{thm3} 
For any $0 < \epsilon < 1$  and any positive integer $n$, 
let $k$ be such that 

%\noindent a)
%\begin{equation}
%\label{jl1a2_c1}
%k \ge \frac{2 \ln n} {\frac{1}{\pi}(2+\epsilon)\epsilon - \ln\left( 2\Phi \left( %\epsilon \sqrt{2/{\pi}}\right)\right)} \text{ .}
%\end{equation}

%\noindent b)
\begin{equation}
\label{jl1a2_c2}
k \ge \frac{2 \ln n}{-\ln (A(s^*))} \text{ .}
\end{equation}
Let $\Gamma$ be a $p$ $\mathbf{x}$ $k$ random matrix with i.i.d. standard Gaussian entries. 
For $\mathbf{x}\in {\mathbf R}^p$, define the mapping $f: {\mathbf R}^p \rightarrow {\mathbf R}^k$ by
$f(\mathbf{x}) = \frac{1}{k}\mathbf{x}\Gamma$. 
Then, for any set $V$ of $n$ points in ${\mathbf R}^p$, 
	such that for any $\mathbf{u}$,$\mathbf{v} \in V$, 
\begin{equation}
\label{rm3a1}
	P\left[(1-\epsilon) \sqrt{\frac{2}{\pi}}\left(\vectornorm{\mathbf{u}-\mathbf{v}}_2\right) \le 
	\vectornorm{\mathbf{f(u)}-\mathbf{f(v)}}_1 \le
	(1+\epsilon) \sqrt{\frac{2}{\pi}}\left(\vectornorm{\mathbf{u}-\mathbf{v}}_2\right)\right ] \ge 1 - \frac{2}{n^{2}} \text{ .}\\
\end{equation}
\end{thm}

%The lower bound for $k$ given in \eqref{jl1a2_c1} is larger than the lower bound for $k$ given in \eqref{jl1a2_c2} since $A_2(s^*)\le A_2(s)$ for all $s>0$ (details in the proof). However, the lower bound for $k$ in \eqref{jl1a2_c1} can be easily computed as opposed to the lower bound for $k$ in \eqref{jl1a2_c2}.

A numerical comparison between the bounds obtained by Matousek \citep{Matousek} and Ailon and Chazelle \citep{Ailon} and the bound given by Theorem \ref{thm3} is presented in table \ref{table:beta4} and fully discussed in section 5.

%%%%%%%%%%%%%%%%%%%%%%%%%
\subsection{$L_2$-$L_1$ norm with Achlioptas-typed Random Matrix}
The following Corollary provides an extension to Theorem \ref{thm3} to the case where the entries of $\Gamma$ are drawn from the Achlioptas types of distribution (eq. \eqref{ach3} with $q=1,2,3$). 

\setcounter{cor}{1}
\begin{cor}\label{cor1}
For any $0 < \epsilon < 1$ and any positive integer $n$, 
let $k$ be as in eq. \eqref{jl1a2_c2} of Theorem \ref{thm3}.
Let $\Gamma$ be a $p$ $\mathbf{x}$ $k$ random matrix with i.i.d. entries drawn from one of Achlioptas distributions (eq. \eqref{ach3} with $q=1,2$ or $3$). 
For $\mathbf{x}\in {\mathbf R}^p$, define the mapping $f: {\mathbf R}^p \rightarrow {\mathbf R}^k$ by
$f(\mathbf{x}) = \frac{1}{k}\mathbf{x}\Gamma$. 
Then, for any set $V$ of $n$ points in ${\mathbf R}^p$, 
	such that for any $\mathbf{u}$,$\mathbf{v} \in V$, 
\begin{equation}
\label{rm3a1}
	P\left[(1-\epsilon) \sqrt{\frac{2}{\pi}}\left(\vectornorm{\mathbf{u}-\mathbf{v}}_2\right) \le 
	\vectornorm{\mathbf{f(u)}-\mathbf{f(v)}}_1 \le
	(1+\epsilon)  \sqrt{\frac{2}{\pi}}\left(\vectornorm{\mathbf{u}-\mathbf{v}}_2\right)\right ] \ge 1 - \frac{2}{n^2} \text{ .}
\end{equation}
\end{cor} 

Note that the lower bound for $k$ using Achlioptas-typed random matrix is the same as the lower bound for $k$ using Gaussian random matrix. The proof of Corollary \ref{cor1} follows from Theorem \ref{thm3} after bounding the moment generating function (mgf) of a Achlioptas-typed random variable by the mgf of a standard Gaussian random variable.

%%%%%%%%%%%%%%%%%%%%%%%%%%%%%%%%%%%%%%%%%%%%%
\section{Concluding remarks} %By working directly with the distributions of the random Euclidean distances rather than the moment generating function, an improvement on the lower bound for $k$ provided by the JL Lemma is obtained when the random matrix consists of i.i.d. standard Gaussian entries ($L_2-L_2$ Random Projection).  

All the results considered in the paper were given in terms of the probability that the distance between one pair of points is not substantially distorted when projected, and a lower bound on this probability was chosen as $1-2/n^2$. However, in most applications, the user is interested in simultaneously preserving distances among all ${n}\choose{2}$ pairs of distinct points selected from $V$. Thus, of interest is a lower bound on the probability of the event
\begin{equation}
\label{union_event}
\Big\{ \underset{\underset{\mathbf{u}\ne \mathbf{v}}{\mathbf{u},\mathbf{v}\in V}}{\bigcap}  (1-\epsilon) \vectornorm{\mathbf{u}-\mathbf{v}}_2 \le \vectornorm{\mathbf{f(u)}-\mathbf{f(v)}}_2 \le (1+\epsilon) \vectornorm{\mathbf{u}-\mathbf{v}}_2  \Big\}
\end{equation}
for example. Since the probability of this event is bounded below by
\begin{equation*}
1-\underset{\underset{\mathbf{u}\ne \mathbf{v}}{\mathbf{u},\mathbf{v}\in V}}{\sum} P\left[ \big\{ (1-\epsilon) \vectornorm{\mathbf{u}-\mathbf{v}}_2 \le \vectornorm{\mathbf{f(u)}-\mathbf{f(v)}}_2 \le (1+\epsilon) \vectornorm{\mathbf{u}-\mathbf{v}}_2\big\}^c\right]
\end{equation*}
where $A^c$ denotes the complement of $A$, and since each term in the sum is less than $2/n^2$, then the probability of the event in \eqref{union_event} is bounded from below by $1/n$. It follows that to obtain a better lower bound for the probability of the event in  \eqref{union_event} using the present techniques, a different bound for the probabilities of the event $\big\{\vectornorm{\mathbf{f(u)}-\mathbf{f(v)}}_2 \ge (1+\epsilon) \vectornorm{\mathbf{u}-\mathbf{v}}_2\big\}$  and
$\big\{\vectornorm{\mathbf{f(u)}-\mathbf{f(v)}}_2 < (1-\epsilon) \vectornorm{\mathbf{u}-\mathbf{v}}_2\big\}$  must be selected. Thus, Achlioptas \citep{Achlioptas} introduces a parameter $\beta>0$ so that for each pair $\mathbf{u}, \mathbf{v} \in V$, 
\begin{equation*}
P\left[(1-\epsilon) \left(\vectornorm{\mathbf{u}-\mathbf{v}}_2\right) \le 
	\vectornorm{\mathbf{f(u)}-\mathbf{f(v)}}_2 \le
	(1+\epsilon)  \left(\vectornorm{\mathbf{u}-\mathbf{v}}_2\right)\right ] \ge 1 - 2/n^{2+\beta} \text{.} 
\end{equation*}
With this choice, the probability of the event in \eqref{union_event} is then seen to be bounded from below by $1-1/n^{\beta}$. The parameter $\beta$ becomes a fine-tuning parameter that affects the probability of the event in \eqref{union_event}. Taking the $\beta>0$ into account in our results, the new expression for the lower bounds for $k$ are as follows:
\begin{itemize}
\item[(i)] Dasgupta and Gupta \citep{Dasgupta}: $k \ge \frac{(24+12\beta)\ln n}{3\epsilon^2-2\epsilon^3}$

\item[(ii)] Theorem \ref{thm1}: $k$ is the smallest even integer satisfying 
$\left(\frac{1+\epsilon}{\epsilon}\right) g(k,\epsilon ) \le \frac{1}{n^{2+\beta}}$, where $g(k,\epsilon)$ is defined in Lemma \ref{lem_g}.

\item[(iii)] Matousek \citep{Matousek}, and Ailon and Chazelle \citep{Ailon}: $k\ge C\epsilon^{-2}\left((4+2\beta)\ln n\right)$

\item[(iv)] Theorem \ref{thm3} and Corollary \ref{cor1}:  $k \ge \frac{(2+\beta) \ln n}{-\ln (A(s^*))}$, where $A(s^*)$ is defined in section 4.
\end{itemize}

The following tables provide a comparison between the results presented here and the results available in the literature.
Table \ref{table:beta} gives a comparison of the lower bounds for $k$ for various values of $n$, $\epsilon$ and $\beta$ obtained from various approaches: Theorem \ref{thm1}, Dasgupta and Gupta's version of the JL Lemma, and exact solution method. The exact solution method numerically finds the smallest integer $k$ such that the sum of the left- and right-tail probabilities, i.e. $P[\vectornorm{\mathbf{y}}^2 \ge \alpha_1 ]+P[\vectornorm{\mathbf{y}}^2 \le \alpha_2 ]$, is less than or equal to $2/n^{2+\beta}$. Note that the exact solution method uses directly the sum of the left- and right-tail probabilities, whereas Theorem \ref{thm1} provides an intermediate bound for the sum of the tail probabilities and then sets the intermediate bound less than or equal to $2/n^{2+\beta}$ to obtain the lower bound for $k$.
The random matrix has i.i.d. standard Gaussian entries. 
We see that the lower bound for $k$ using Theorem \ref{thm1} 
is very close to the lower bound for $k$ using the exact solution method, and significantly improves on the lower bound for $k$ given by Dasgupta and Gupta's version of the JL Lemma. The advantage provided by our approach is reflected in the additional percentage dimension reduction of at least $13\%$ in all cases considered. In some of the cases, we achieve a $30\%$ additional reduction in dimension when compared to the Dasgupta and Gupta bound.

Table \ref{table:beta4} compares the lower bound for $k$ obtained from Ailon and Chazelle \citep{Ailon} and Matousek \citep{Matousek} for $L_2$-$L_1$ distance ($C=1$), and Theorem \ref{thm3} for $L_2$-$L_1$ distance. The random matrix has 
i.i.d. standard Gaussian entries. We observe that the lower bounds for $k$ from Ailon and Chazelle \citep{Ailon} and Matousek \citep{Matousek} are significantly  larger  than the lower bound for $k$ obtained from Theorem \ref{thm3}. In most cases, the results of Theorem \ref{thm3} provide an additional reduction of $35\%-40\%$ in the lower bound for $k$.

\begin{table}[!ht]
\normalsize\caption{\normalsize{Comparison of the lower bounds for $k$  for $L_2$-$L_2$ distance: exact solution (numerically solving for $k$ after setting the sum of left and right-tail probabilities equal to $2/n^{2+\beta}$), Theorem \ref{thm1}, and JL Lemma.}}
\label{table:beta}
\begin{center}
\begin{tabular}{|rr|r|r|r|r|}
\hline
N(0,1) entries &            & {\bf exact solution} & {\bf Theorem \ref{thm1}} &   {\bf JL Lemma} \\

       n=50 &   $\epsilon$ = .1, $\beta=1$ &        3776 &                3976 &       5030 \\

           &   $\epsilon$ = .3, $\beta=1$&         456 &                     494 &        653 \\

           &   $\epsilon$ = .1, $\beta=2$ &         5336 &                        5572 &         6707 \\
           &   $\epsilon$ = .3, $\beta=2$ &         654 &                        692 &         870 \\
\hline
       n=100 &   $\epsilon$ = .1, $\beta=1$ &        4601 &                4822 &       5921 \\

           &   $\epsilon$ = .3, $\beta=1$&         561 &                     598 &        768 \\

           &   $\epsilon$ = .1, $\beta=2$ &         6461 &                        6716 &         7895 \\
           &   $\epsilon$ = .3, $\beta=2$ &         797 &                        834 &         1024 \\
\hline
       n=500 &   $\epsilon$ = .1, $\beta=1$ &        6552 &                6808 &       7991 \\

           &   $\epsilon$ = .3, $\beta=1$&         808 &                     846 &        1036 \\

           &   $\epsilon$ = .1, $\beta=2$ &         9110 &                        9390 &         10654 \\
           &   $\epsilon$ = .3, $\beta=2$ &         1130 &                        1168 &         1382 \\
\hline
       n=1000 &   $\epsilon$ = .1, $\beta=1$ &        7403 &                7670 &       8882 \\

           &   $\epsilon$ = .3, $\beta=1$&         916 &                     954 &        1152 \\

           &   $\epsilon$ = .1, $\beta=2$ &         10262 &                        10548 &         11842 \\
           &   $\epsilon$ = .3, $\beta=2$ &         1274 &                        1312 &         1536 \\
\hline
\end{tabular}
\end{center}  
\end{table}   

\begin{table}[!ht]
%\vspace{-4pc}
\normalsize\caption{\normalsize{Normal random matrix: comparison of the lower bounds for $k$ from Matousek\text{ }\citep{Matousek} for $L_2$-$L_1$ distance ($C=1$), and Theorem \ref{thm3} for $L_2$-$L_1$ distance.}}
\label{table:beta4}
\begin{center}
\begin{tabular}{|rr|r|r|r|}
\hline
N(0,1) entries  &         & {\bf $L_2-L_1$ Matousek} &  {\bf $L_2-L_1$ Theorem \ref{thm3}}\\
\hline
       n=50 &   $\epsilon$ = .1, $\beta=1$ &     2348 &        1398\\

           &   $\epsilon$ = .3, $\beta=1$ &        261 &         168\\

           &   $\epsilon$ = .1, $\beta=2$ &        3130 &          1863\\

           &   $\epsilon$ = .1, $\beta=2$ &        348 &          223\\
\hline
       n=100 &   $\epsilon$ = .1, $\beta=1$ &     2764 &        1645\\

           &   $\epsilon$ = .3, $\beta=1$ &        308 &         197\\

           &   $\epsilon$ = .1, $\beta=2$ &        3685 &          2193\\

           &   $\epsilon$ = .1, $\beta=2$ &        410 &          263\\
\hline
       n=500 &   $\epsilon$ = .1, $\beta=1$ &     3729 &        2220\\

           &   $\epsilon$ = .3, $\beta=1$ &        415 &         266\\

           &   $\epsilon$ = .1, $\beta=2$ &        4972 &          2960\\

           &   $\epsilon$ = .1, $\beta=2$ &        553 &          354\\
\hline
       n=1000 &   $\epsilon$ = .1, $\beta=1$ &     4145 &        2468\\

           &   $\epsilon$ = .3, $\beta=1$ &        461 &         296\\

           &   $\epsilon$ = .1, $\beta=2$ &        5527 &          3290\\

           &   $\epsilon$ = .1, $\beta=2$ &        615 &          394\\
\hline
\end{tabular}  
\end{center}
\end{table}

\clearpage

%------------------------------------------%
\appendix
\section{Proofs}
%\noindent\textbf{Proof of Theorem \ref{thm1}:} It is easy to see that the equality in \eqref{gamma1} is
%obtained by repeated integration by parts.\\
\noindent\textbf{Proof of Lemma \ref{lem_g}:} 
Proving that $g(k+2,\epsilon) \le g(k,\epsilon)$ is equivalent to proving that
\begin{equation}
(1+\epsilon) e^{-(1+\epsilon)} \left(1+\frac{1}{d}\right)^{d} \le 1\text{ .}
\end{equation}
Observe that $\left(1+\frac{1}{d}\right)^{d} \le e$, and thus, 
\begin{equation}
(1+\epsilon) e^{-(1+\epsilon)} \left(1+\frac{1}{d}\right)^{d} \le (1+\epsilon) e^{-\epsilon} \le 1\text{ .}
\end{equation}

\noindent\textbf{Proof of Theorem \ref{thm1} Part a:} Suppose $1 < d < \lambda_1$. Dividing both sides of 
\eqref{rm2ob} by $\left(\frac{\lambda_1}{\lambda_1 -d}\right)
\left(\frac{\lambda_1^{d-1}}{(d-1)!} \right)$, it is seen that \eqref{rm2ob} is equivalent to
\begin{equation}
\label{rm2oe}
\frac{\lambda_1 - d}{\lambda_1} \left( 1 + \frac{d-1}{\lambda_1} + 
	\frac{(d-1)(d-2)}{\lambda_1^2} + \dots + \frac{(d-1)!}{\lambda_1^{d-1}} \right) 
\le 1  \text { .}
\end{equation}
But 
\begin{eqnarray}
\label{rm2of1}
1 + \frac{d-1}{\lambda_1} + 
	\frac{(d-1)(d-2)}{\lambda_1^2} + \dots + \frac{(d-1)!}{\lambda_1^{d-1}} 
&\le \sum_{i=0}^{d-1} \left(\frac{d-1}{\lambda_1}\right)^i  \\
\label{rm2of2}
&\le \sum_{i=0}^{d-1} \left(\frac{d}{\lambda_1}\right)^i  \\
\label{rm2of3}
&= \frac{1 - \left(\frac{d}{\lambda_1}\right) ^d}{1 - \frac{d}{\lambda_1}}
\end{eqnarray} 
where \eqref{rm2of3} is obtained from the finite geometric sum.

The inequality in \eqref{rm2oe} follows immediately from \eqref{rm2of3}.

\noindent\textbf{Proof of Theorem \ref{thm1} Part b:} Suppose $0 < \lambda_2 < d$. Dividing both sides of 
\eqref{rm2od} by $\left(\frac{\lambda_2}{d - \lambda_2}\right)
\left(\frac{\lambda_2^{d-1}}{(d-1)!} \right)$,  \eqref{rm2od} is seen to be equivalent to
\begin{equation}
\label{rm2oh}
\frac{d - \lambda_2}{\lambda_2} \frac{\lambda_2}{d} \left( 1 + \frac{\lambda_2}{d+1} + 
	\frac{\lambda_2^2}{(d+1)(d+2)} + \dots \right) 
\le 1  \text{ .}
\end{equation}

But, 
\begin{eqnarray}
\label{rm2oi1}
1 + \frac{\lambda_2}{d+1} + 
	\frac{\lambda_2^2}{(d+1)(d+2)} + \dots 
&\le \sum_{i=0}^{\infty} \left(\frac{\lambda_2}{d+1}\right)^i\\
&\le \sum_{i=0}^{\infty} \left(\frac{\lambda_2}{d}\right)^i\\ 
\label{rm2oi2}
&= \frac{d}{d-\lambda_2} \text{ .}
\end{eqnarray}
Thus, \eqref{rm2oh} follows immediately from  \eqref{rm2oi2}.

%%%%%%%%%%%%%%%%%%%%%%%%%%%%%%%%%%%%
\noindent\textbf{Proof of Theorem \ref{thm3}:} Let $\Gamma$ be a random matrix of 
	dimension $p$ x $k$ with i.i.d entries $r_{ij} \sim N(0,1)$. For $\mathbf{x}\in \mathbf{R}^p$, define a linear mapping $f\colon{\mathbf R}^p 
	\rightarrow {\mathbf R}^k$ by $f(\mathbf{x}) = \frac{1}{k}\mathbf{x}\Gamma$. Let 
\begin{equation}
\label{rm2op1}
y_j = \frac{\mathbf{x}r_j}{\vectornorm{\mathbf{x}}_2} \sim N(0,1) \text{ .} 
\end{equation}
Then, $E(\vectornorm{\mathbf{y}}_1)=k\sqrt{2/{\pi}}$, and $M_{|y_j|}(s) = 2e^{s^2/2}\Phi(s)$.  

Let $\alpha_1 = k\sqrt{2/\pi} (1+\epsilon)$, then the right-tail probability is bounded by
 \begin{align}
	P\left[\vectornorm{\mathbf{f(x)}}_1 \ge \sqrt{2/\pi} (1+\epsilon ) \vectornorm{\mathbf{x}}_2\right ]
	&= P\left[\vectornorm{\mathbf{y}}_1 \ge  \alpha_1 \right] \nonumber\\
\label{rm3aa2b}
	&\le \left(2e^{-(s\alpha_1/k) + (s^2/2)} \Phi(s) \right)^k \text{ ,}\quad s>0.
\end{align}
Let  $A(s) = e^{-(s\alpha_1/k) + (s^2/2)} \Phi(s)$, and denote by $s^*$ the minimizer of $A$, so that $s^*$ is the solution to
\begin{equation}
\label{rm3a4ib}
s = \sqrt{2/\pi} (1+\epsilon) - \frac{\phi (s)}{\Phi (s)} \text{ .}
\end{equation}
The second derivative of $A(s)$ with respect of $s$ is taken to ensure that $s^*$ is the minimizer of $A$. 
\begin{equation}
\label{rm3a4ib3}
A''(s) = e^{-(s\alpha_1/k) + (s^2/2)} \left[\left(\left( s-\frac{\alpha_1}{k}\right)^2+1\right)\Phi(s) +\left(s-2\frac{\alpha_1}{k}\right)\phi(s)\right] \text{ .}
\end{equation}
Note that  for $s>0$,
\begin{equation}
\left( s-\frac{\alpha_1}{k}\right)^2+1 > 2\left(\frac{\alpha_1}{k}\right) \left(\frac{\phi(s)}{\Phi(s)}\right) \text{ .}
\end{equation}
Thus, $A''(s) > 0$, which implies $s^*$ is the unique minimizer of $A$. 
Setting $A(s^*)\le 1/{n^2}$, we obtain the lower bound for $k$ to be $k \ge \frac{2\ln n}{-\ln A(s^*)}$.

Similarly, let $\alpha_2 = k\sqrt{2/\pi} (1-\epsilon)$, then left-tail probability is bounded by
\begin{align}
P\left[\vectornorm{\mathbf{f(x)}}_1 \le \sqrt{2/{\pi}} (1-\epsilon ) \vectornorm{\mathbf{x}}_2\right ]
&= P[\vectornorm{\mathbf{y}}_1 \le \alpha_2 ] \nonumber\\
\label{rm3a4dl2}
	&\le \left( 2e^{(s\alpha_2/k) +(s^2/2)} \left(1-\Phi(s) \right)\right)^k \text{ ,}\quad s>0 \text{ .}
%\label{rm3a4ic3}
%	&\le \left( 2e^{-(s\alpha_1/k) +(s^2/2)} \Phi(s)\right)^k \text{ .}	
\end{align}
Let 
\begin{equation}
B(s) = 2e^{(s\alpha_2/k) +(s^2/2)} \left(1-\Phi(s) \right) \text{ .}
\end{equation}
The next proposition provides $B(s) \le A(s)$. 

\begin{prop}\label{prop4} For all $\zeta>0$, we have
\begin{equation}
\label{rm3a4gc1}
e^{2\sqrt{2/\pi} \zeta} < \frac{\Phi (\zeta)}{1-\Phi (\zeta)} \text{ .}
\end{equation} 
\end{prop}
\noindent\textbf{Proof of Proposition \ref{prop4}:} Let $f(\zeta) = \frac{\Phi (\zeta)}{1-\Phi (\zeta)} e^{-2\zeta\sqrt{2/\pi}}$, then eq. \eqref{rm3a4gc1} is equivalent to
\begin{equation}
\label{rm3a4gc1b}
 f(\zeta) > 1
\end{equation}
It suffices to prove that $f(\zeta)$ is an increasing function. Taking the derivative of $f$ with respect to $\zeta$ yields
\begin{equation}
\label{rm3a4gc1c}
 f'(\zeta) = \frac{e^{-2\zeta\sqrt{2/\pi}}}{1-\Phi(\zeta)} \left[\frac{\phi(\zeta)}{1-\Phi(\zeta)} - 2\sqrt{2/{\pi}} \Phi(\zeta) \right] \text{ .}
\end{equation}
We should note that the first term is positive. The ratio $\frac{\phi(\zeta)}{1-\Phi(\zeta)}$ is the inverse of the Mill's ratio, which is an increasing function, and we observe that
\begin{equation}
\label{rm3a4gc1d}
\frac{\phi(\zeta)}{1-\Phi(\zeta)} > 2\sqrt{2/{\pi}} \Phi(\zeta)
\end{equation}
which implies $f'(\zeta)>0$, and hence, $f$ is an increasing function of $\zeta$. The minimum of $f$ is attained when $\zeta=0$. In other words, $\underset{\zeta}{\min} f(\zeta) = 1$, and hence eq. \eqref{rm3a4gc1b} is proven. 

Using Proposition \ref{prop4} with $\zeta=s$, $B(s) \le A(s)$ for $s>0$. Thus, the left-tail probability is bounded by
\begin{equation}
P\left[\vectornorm{\mathbf{f(x)}}_1 \le \sqrt{2/{\pi}} (1-\epsilon ) \vectornorm{\mathbf{x}}_2\right ]
\label{rm3a4ic3}
	\le \left( 2e^{-(s\alpha_1/k) +(s^2/2)} \Phi(s)\right)^k \text{ .}	
\end{equation}

Note that the right side of inequality \eqref{rm3a4ic3} for the left-tail probability is the same as in the case for the right-tail probability.

%%%%%%%%%%%%%%%%%%%%%%%%%%%%%%%%%%%%%%%
\noindent\textbf{Proof of Corollary \ref{cor1}:} Let $\Gamma$ be a random matrix of 
	dimension $p$ x $k$ with i.i.d entries from an Achlioptas distribution ($q=1, 2$ or $3$). For $\mathbf{x}\in \mathbf{R}^p$, define a linear mapping $f\colon{\mathbf R}^p 
	\rightarrow {\mathbf R}^k$ by $f(\mathbf{x}) = \frac{1}{k}\mathbf{x}\Gamma$. Let 
\begin{equation}
\label{rm2op1}
y_j = \frac{\mathbf{x}r_j}{\vectornorm{\mathbf{x}}_2} = \sum_{i=1}^p c_i r_{ij}\text{ ,} 
\end{equation}
where $c_i = \frac{x_i}{\vectornorm{\mathbf{x}}_2}$, so that $\sum_{i=1}^p c_i^2=1$. 
Then, $E(\vectornorm{\mathbf{y}}_1)=k\sqrt{2/{\pi}}$, and 
\begin{equation}
\label{ach_myj}
M_{y_j}(t) =  \prod_{i=1}^p \left(1+\frac{1}{q}\left(\cosh(c_it\sqrt{q})-1\right)\right)  \text{ ,  } \forall t \text{ .}   
\end{equation}

We introduce the following proposition to provide a bound on $M_{y_j}(t)$. 
\begin{prop} \label{prop2}For $x\in \mathbf{R}$, and $q=1,2$ or $3$, we have
\begin{equation}
\label{rm2r1d3a3}
1+ \frac{1}{q}\left(\cosh \left( x\sqrt{q} \right) -1\right) \le e^{x^2/2}
\end{equation}
\end{prop}
\noindent \textbf{Proof of Proposition \ref{prop2}: } 
Our proof will show that $g(x)=\frac{\log (\cosh (x))}{x^2/2}$ takes as its maximum value of $1$ at $x=0$. By a symmetry argument, we only need to consider the case $x>0$ and show that $g$ is decreasing in $x>0$. 

For $q=1$, $g(x)=\frac{\log (\cosh (x))}{x^2/2}$ is decreasing in $x>0$. 

For the cases $q=2$ and $q=3$, $g(x)=\frac{1+ \frac{1}{q}\left(\cosh \left( x\sqrt{q} \right) -1\right)} {e^{x^2/2}}$. 
To prove that $g$ is decreasing, we need $g'(x) < 0$. 
\begin{equation}
\label{rm2r1d3a5}
g'(x)=\frac{1}{e^{x^2/2}} \left( \frac{\sqrt{q}}{q}\sinh (x\sqrt{q}) - x\left( 1+\frac{1}{q}(\cosh (x\sqrt{q}) - 1) \right)  \right)
\end{equation}
with $g'(0)=0$. Let 
\begin{equation}
\label{rm2r1d3a5}
h(x)= \frac{\sqrt{q}}{q}\sinh (x\sqrt{q}) - x\left( 1+\frac{1}{q}(\cosh (x\sqrt{q}) - 1) \right)
\end{equation}
 Since $x>0$, and $h(0)=0$, if $h'(x)<0$, then $x=0$ is maximum and $h(x)<0$. But
\begin{equation}
\label{rm2r1d3a5}
h'(x)= \left(\frac{q-1}{q}\right) ( \cosh (x\sqrt{q})-1) - x\frac{\sqrt{q}}{q}\sinh (x\sqrt{q})
\end{equation}
with $h'(0)=0$. Let
\begin{equation}
\label{rm2r1d3a6}
l(x)= (q-1)(\cosh (x\sqrt{q})-1) - x\sqrt{q}\sinh (x\sqrt{q})
\end{equation}
then
\begin{equation}
\label{rm2r1d3a6}
l'(x)= \sqrt{q}(q-2)\sinh (x\sqrt{q}) - x q\cosh (x\sqrt{q})
\end{equation}
with $l'(0)=0$.
For $q=2$, we have $l'(x) = - 2 x \cosh (x\sqrt{2}) < 0$, which implies $g(x)$ is decreasing for $x>0$. 

For $q=3$, let
\begin{equation}
\label{rm2r1d3a7}
m(x)=l'(x)= \sqrt{3}\sinh (x\sqrt{3}) - 3 x\cosh (x\sqrt{3})
\end{equation}
then $m'(x)= - 3 \sqrt{3} \sinh (x\sqrt{3}) < 0$, which implies $g(x)$ is decreasing for $x>0$. Thus, Proposition \ref{prop2} is proven. \hspace{19pc}$\Box$

\vspace{1.3pc}

Using Proposition \ref{prop2}, for $t \in \mathbf{R}$, $q=1, 2$ or $3$, and $Z\sim N(0,1)$,
\begin{equation}
\label{rm2r1d3b1}
M_{y_{j}} (t) = \prod_{i=1}^p \left(1+ \frac{1}{q}\left(\cosh \left( c_i t\sqrt{q} \right) -1\right) \right) \le \prod_{i=1}^p e^{c_i^2t^2/2} = e^{t^2/2} = M_{Z}(t) \text{ .}
\end{equation}
The inequality in \eqref{rm2r1d3b1} implies
\begin{equation}
\label{rm2r1d3b2}
M_{|y_{j}|} (t) \le M_{|Z|}(t) = 2e^{t^2/2}\Phi(t) \text{ , } t\in \mathbf{R} \text{ .}
\end{equation}
Thus, for $\alpha_1 = k\sqrt{2/\pi} (1+\epsilon)$, the right-tail probability is bounded by
 \begin{align}
	P\left[\vectornorm{\mathbf{f(x)}}_1 \ge \sqrt{2/\pi} (1+\epsilon ) \vectornorm{\mathbf{x}}_2\right ]
	&= P\left[\vectornorm{\mathbf{y}}_1 \ge  \alpha_1 \right] \nonumber\\
	&\le \left(2e^{-(s\alpha_1/k)} M_{|y_j|}(s) \right)^k \text{ ,}\quad s>0\nonumber\\
	&\le \left(2e^{-(s\alpha_1/k)} M_{|Z|}(s) \right)^k \nonumber\\
\label{rm3bb2b}
	&\le \left(2e^{-(s\alpha_1/k) + (s^2/2)} \Phi(s) \right)^k
\end{align}
where the last inequality \eqref{rm3bb2b} is the same as in the case of Gaussian random matrix.

Similarly, for $\alpha_2 = k\sqrt{2/\pi} (1-\epsilon)$, the left-tail probability is bounded by
 \begin{align}
	P\left[\vectornorm{\mathbf{f(x)}}_1 \le \sqrt{2/\pi} (1-\epsilon ) \vectornorm{\mathbf{x}}_2\right ]
	&= P\left[\vectornorm{\mathbf{y}}_1 \le  \alpha_2 \right] \nonumber\\
	&\le \left(2e^{(s\alpha_2/k)} M_{|y_j|}(-s) \right)^k \text{ ,}\quad s>0\nonumber\\
	&\le \left(2e^{(s\alpha_2/k)} M_{|Z|}(-s) \right)^k \nonumber\\
\label{rm3bb2bc}
	&\le \left(2e^{(s\alpha_2/k) + (s^2/2)} \left(1-\Phi(s)\right) \right)^k\\
	&\le \left(2e^{-(s\alpha_1/k) + (s^2/2)} \Phi(s) \right)^k \text{ .}
\end{align}
The last inequality is the same as in the case of the right-tail probability, and hence, we are done.

\section*{Acknowledgements} Research for this article was partially supported by
NSF Grant SES-0532346, NSA RUSIS Grant H98230-06-1-0099, NSF REU Grant MS-0552590,
and NCI Grant T32CA96520.

\vspace{3pc}

\textbf{Javier Rojo}, Statistics Professor at Rice University, obtained a master's degree from Stanford University and a PhD degree in Statistics from UC Berkeley. He is also Adjunct Professor at MD Anderson Cancer Center, and Director of the Rice Summer Institute of Statistics. He is Editor of the Journal of Nonparametric Statistics, and Chair/Organizer of the Lehmann Symposia. He edited three volumes (IMS-LNMS) and currently edits the Lehmann Collected works. He is ASA, IMS, and AAAS Fellow and an elected member of ISI. He chaired the ASA Committee on Fellows, served as NSF program director, and served in National Academy of Sciences committees.
\vspace{.75pc}

\textbf{Tuan S. Nguyen} is currently a Ph.D. student in the statistics department at Rice University under the direction of Professor Javier Rojo. He obtained a master's degree in Statistics from Rice University in 2009, and a bachelor degree in Engineering Mathematics and Statistics from the University of California at Berkeley in 2004. His research interests include Dimension Reduction, Survival Analysis, Microarray Data Analysis, Data Mining and Clinical Trials.


\begin{thebibliography}{1}
{\normalsize
\bibitem[Achlioptas, 2001]{Achlioptas} Achlioptas, D. Database-friendly random projections. {\em Proc.
 ACM Symp. on the principles of database systems}, 274--281, 2001. 

\bibitem[Achlioptas2, 2001]{Achlioptas2} Achlioptas, D., McSherry, F., and Scholkopf, B. Sampling techniques for kernel methods. In {\em Advances in Neural Information Processing Systems}, 335--342, 2001. 

\bibitem[Ailon and Chazelle, 2006]{Ailon} Ailon, N., and Chazelle, B. Approximate nearest neighbors and the fast Johnson-Lindenstrauss transform. {\em Proc.
 38th ACM Symp. Theory of Computing}, 557--563, 2006. 

\bibitem[Ailon and Liberty, 2008]{Ailon2} Ailon, N., and Liberty, E. Fast dimension reduction using Rademacher series on dual BCH codes. In {\em Symp. on Discrete Algorithms}, 1--9. San Francisco, CA, 2008. 

\bibitem[Ailon \textit{et al.}, 2008]{Ailon3} Ailon, N., Liberty, E., and Singer A. Dense Fast Random Projections and Lean Walsh Transforms. In {\em Proc. of 11th and 12th International Workshop on Approximation, Randomization and Combinatorial Optimization: Algorithms and Techniques}, 512--522. Springer-Verlag, 2008. 

%\bibitem[Amador, 2007]{Amador}, J.J. Random projection and orthonormality for lossy image compression. {\em Image and Vision Computing 25}, 754--766, 2007.

\bibitem[Arriaga and Vempala, 1999] {Arriaga} Arriaga, R.I., and Vempala, S. An algorithmic theory of learning:
 robust concepts and random projections. {\em 40th Annual Symposium on Foundations
 of Computer Science}. New York, NY, 1999.

\bibitem[Bertoni and Valentini, 2005] {Bertoni1} Bertoni, A., and Valentini, G. Random projections for assessing gene expression cluster stability. In {\em IJCNN 2005, the IEEE-INNS International Joint Conference on Neural Networks}. Montreal, 2005.

\bibitem[Bertoni and Valentini, 2006] {Bertoni2} Bertoni, A., and Valentini, G. Ensembles based on random projections to improve the accuracy of clustering algorithms. \url{http://eprints.pascal-network.org/archive/00002362/01/bertoni-vale-WIRN05.pdf}. Milano, 2006.

\bibitem[Bertoni \textit{et al.}, 2008] {Bertoni3} Bertoni, A., Valentini, G., Folgieri, R., and Piuri, V. Ensembles based on random projections for gene expression data analysis. \url{http://www.mtcube.com/Tesi-Folgieri.pdf}. Archivio Istituzionale della Ricerca, Milano, 2008.

\bibitem[Bingham and Mannila, 2001]{Bingham} Bingham, E. and Mannila, H. Random projection in dimensionality reduction: applications to image and text data. In {\em Proc. of KDD}, pp. 245--250, San Francisco, CA, 2001.  

\bibitem[Blum, 2005]{Blum} Blum, A. Random projection, margins, kernels, and feature selection. In {\em SLSFS 2005, LNCS 3940}, pp. 52--68, 2005.  

\bibitem[Brinkman and Charikar, 2003]{Brinkman} Brinkman, B. and Charikar, M. On the impossibility of dimension reduction in $L_1$. {\em Proc. 44th IEEE Symp Foundations of Computer Science}, 514--523, 2003.  

\bibitem[Buhler and Tompa, 2002]{Buhler} Buhler, J. and Tompa M. Finding motifs using random projections. {\em Journal of Computational Biology}, \textbf{9.2}, 225--242, 2002.  

\bibitem[Candes and Tao, 2006]{Candes} Candes E.J., and Tao, T. Near-optimal signal recovery from random projections: universal encoding strategies?.
{\em Information Theory, IEEE Transactions on, 52.12}, 5406--5425, 2006.

%\bibitem[Casella and Berger, 2002]{CB} Casella G., and Berger, R.L. Statistical Inference, Second Edition.
%{\em Duxbury Advanced Series}, 2002.

\bibitem[Charikar and Sahai, 2002]{Charikar} Charikar M., and Sahai, A. Dimension reduction in $L_1$ norm.
In {\em Proceedings of the 43rd Annual IEEE Symp. on Foundations of Computer Science}, 551--560, 2002.

\bibitem[Dasgupta and Gupta, 1999]{Dasgupta} Dasgupta, S. and Gupta, A. An elementary proof of the Johnson-Lindenstrauss lemma. {\em Random Structures and Algorithms 22.1}, 60--65, 2003.

\bibitem[Dasgupta, 2000]{Dasgupta2} Dasgupta, S. Experiments with random projection. In {\em Uncertainty in Artificial Intelligence}, 2000.

\bibitem[Deegalla and Bostrum, 2006]{Deegalla} Deegalla, S., and Bostrum, H. Reducing high-dimensional data by principal component analysis vs. random projection for nearest neighbor classification. In {\em Proc. of the 5th International Conference on Machine Learning and Applications}, 245--250, 2006.

\bibitem[Fern and Brodley, 2003]{FB} Fern, X.Z. and Brodley, C.E. Random projection for high dimensional data clustering: A cluster ensemble approach. In {\em Proc. of the Twentieth International Conference on Machine Learning}, 2003.

\bibitem[Fradkin and Madigan, 2002]{Fradkin} Fradkin, D. and Madigan, D. Experiments with 
Random Projections for Machine Learning. {\em ACM}, 2002.

\bibitem[Frankl and Maehara, 1988]{Frankl1} Frankl, P. and Maehara, H. The Johnson-Lindenstrauss lemma 
and the sphericity of some graphs. {\em J. Combin. Theory Ser. B44(3)}, 355--362, 1988.

\bibitem[Goel, 2005]{Goel} Goel, N., Bebis G., and Nefian A. Face recognition experiments with random projection. {\em Proc. SPIE}, \textbf{5779}, pp. 426--437, 2005. doi:10.1117/12.605553.  

\bibitem[Hecht-Nielsen, 1994]{Hecht_Nielsen} Hecht-Nielsen, R. Context vectors: General purpose approximate meaning representations self-organized from raw data. In {\em Computational Intelligence: Imitating Life (Zurada et al. eds.)}, pp. 43--56, 1994. 

\bibitem[Indyk, 2006]{Indyk1} Indyk, P. Stable distributions, pseudorandom generators,
embeddings, and data stream computation. In {\em Journal of the ACM 53.3}, 307--323, 2006.

\bibitem[Indyk and Motwani, 1998]{Indyk2} Indyk, P., and Motwani, R. Appropriate nearest neighbors: towards removing the curse of dimensionality. In {\em Proc. 30th ACM Symp. on Theory of Computing}, 604--613, 1998.

%\bibitem[Indyk, 2001]{Indyk} Indyk, P. Algorithmic applications in low-distortion embeddings. In {\em Proc. 42nd IEEE Symp Foundations of Computer Science}, 10--35, 2001.

\bibitem[Johnson and Lindenstrauss, 1984]{JL} Johnson, W. and Lindenstrauss, J. Extensions of Lipschitz maps into a 
Hilbert space. {\em Contemp. Math. 26}, 189--206, 1984.

\bibitem[Kaski, 1998]{Kaski} Kaski, S. Dimensionality reduction by random mapping: Fast similarity computation for clustering. In {\em Proc. of IJCNN 26}, 413--418, Piscataway, NJ, 1998.

\bibitem[Kleinberg, 1997]{Kleinberg} Kleinberg, J.M. Two algorithms for nearest-neighbor search in higher dimensions. In {\em Proc. of 29th ACM Symp. on Theory of Computing}, 599--608, 1997.

\bibitem[Kurimo, 1999]{Kurimo} Kurimo, M. Indexing audio documents by using latent semantic analysis and SOM. E. Oja and S. Kaski (eds.), Kohonen Maps, 1999.

\bibitem[Lee and Naor, 2004]{Lee} Lee, J.R. and Naor, A. Embedding the diamond graph in $L_p$ and dimension reduction in $L_1$. {\em Geom Funct. Anal 14}, 745--747, 2004.

\bibitem[Li \textit{et al.}, 2006]{Li4} Li, P., Hastie, T.J., and Church K.W. Very Sparse Random Projections.
{\em Proceedings of the 12th ACM SIGKDD international conference on knowledge
discovery and data mining}, 287--296, 2006. 

\bibitem[Li \textit{et al.}, 2007]{Li5} Li, P., Hastie, T.J., and Church K.W. Nonlinear tail bounds for dimension reduction in $l_1$ using cauchy random projections.
{\em Journal of Machine Learning Research 8}, 2497--2532, 2007. 

\bibitem[Liu \textit{et al.}, 2006]{Liu} Liu, K., Kargupta, H., and Ryan J. Random projection-based multiplicative data perturbation for privacy preserving distributed data mining.
{\em IEEE Transactions on Knowledge and Data Engineering 18.1}, 2006. 

\bibitem[Mannila \textit{et al.}, 2002]{Mannila} Mannila, H., Hollmen, J., Seppanen, J.K.,
Korpiaho, K., Tikanmaki, J., Bingham, E. From Data to Knowledge
Research Unit. {\em Technical Report}, 2002. 

\bibitem[Matousek, 2002]{Matousek2} Matousek, J. Lectures in Discrete Geometry. {\em Springer}, New York, 2002. 

\bibitem[Matousek, 2007]{Matousek} Matousek, J. On variants of the Johnson-Lindenstrauss Lemma. {\em Wiley InterScience}, 2007. doi: 10.1002/rsa.20218. 

\bibitem[Papadimitriou \textit{et al.}, 1998]{Papadimitriou} Papadimitriou, C.H., Raghvan, P., Tamaki, H., and
Vempala, S. Latent semantic analysis: A probabilistic analysis. In {\em Proc. of 17th ACM Symp. On the principles of Database Systems}, pp. 159--168, 1998.

\bibitem[Vempala, 1998]{Vempala} Vempala, S. Random projection: A new approach to VLSI layout. In {\em Proc. of FOCS}, pp. 389--395, Palo Alto, CA, 1998.
 
}

\end{thebibliography}
\end{document}